\documentclass{article}

\usepackage{arxiv}

\usepackage[utf8]{inputenc} 
\usepackage[T1]{fontenc}    
\usepackage{hyperref}       
\usepackage{url}            
\usepackage{booktabs}       
\usepackage{amsfonts}       
\usepackage{nicefrac}       
\usepackage{microtype}      
\usepackage{graphicx}
\usepackage{doi}
\usepackage{graphics} 
\usepackage{epsfig} 
\usepackage{mathptmx} 
\usepackage{times} 
\usepackage{amsmath} 
\usepackage{amsmath} 
\usepackage{amssymb}  
\usepackage{algorithm}
\usepackage{algorithmic}
\usepackage{subcaption}
\usepackage[dvipsnames]{xcolor}

\usepackage{pgfplots} 
\pgfplotsset{width=1\linewidth,compat=1.9}

\title{Privacy-preserving and Uncertainty-aware Federated Trajectory Prediction for Connected Autonomous Vehicles}

\author{
        Muzi Peng\\
	Department of Electrical and Computer Engineering\\
	Northeastern University\\
	Boston, MA 02115 \\
	\texttt{peng.mu@northeastern.edu} \\
	\And
	Jiangwei Wang\\
	Department of Electrical and Computer Engineering\\
	University of Connecticut\\
	Storrs, CT 06268\\
	\texttt{jiangwei.wang@uconn.edu} \\
        \And
	Dongjin Song\\
	Department of Computer Science and Engineering\\
	University of Connecticut\\
	Storrs, CT 06268\\
	\texttt{dongjin.song@uconn.edu} \\
        \And
	Fei Miao\\
	Department of Computer Science and Engineering\\
	University of Connecticut\\
	Storrs, CT 06268\\
	\texttt{fei.miao@uconn.edu} \\
        \And
	Lili Su\\
	Department of Electrical and Computer Engineering\\
	Northeastern University\\
	Boston, MA 02115 \\
	\texttt{l.su@northeastern.edu} \\
}

\renewcommand{\shorttitle}{Privacy-preserving and Uncertainty-aware Federated Trajectory Prediction for CAVs}

\hypersetup{
pdftitle={Privacy-preserving and Uncertainty-aware Federated Trajectory Prediction for Connected Autonomous Vehicles},
pdfsubject={q-bio.NC, q-bio.QM},
pdfauthor={Muzi Peng, Jiangwei Wang, Dongjin Song, Fei Miao, Lili Su},
}

\begin{document}
\maketitle

\begin{abstract}
    Deep learning is the method of choice for trajectory prediction for autonomous vehicles. Unfortunately, its data-hungry nature implicitly requires the availability of sufficiently rich and high-quality centralized datasets, which easily leads to privacy leakage. 
    Besides, uncertainty-awareness becomes increasingly important for safety-crucial cyber physical systems whose prediction module  heavily relies on machine learning tools.    
    In this paper, we relax the data collection requirement and enhance uncertainty-awareness 
    by using Federated Learning on Connected Autonomous Vehicles with an uncertainty-aware global objective. We name our algorithm as FLTP. 
    We further introduce ALFLTP which boosts FLTP via using active learning techniques in 
    adaptatively selecting participating clients. 
    We consider two different metrics negative log-likelihood (NLL) and aleatoric uncertainty (AU) for client selection. Experiments on Argoverse dataset show that FLTP significantly outperforms the model trained on local data. In addition, ALFLTP-AU converges faster in training regression loss and performs better in terms of NLL, minADE and MR than FLTP in most rounds, and has more stable round-wise performance than ALFLTP-NLL.
\end{abstract}


\section{Introduction}
\label{sec: intro}
Accurate trajectory prediction of surrounding objects is crucial for autonomous driving. 
For example, it is important to predict the lane merging or overtaking actions of neighboring vehicles in order to avoid collision. 
Recently, deep learning has been the method of choice for trajectory prediction \cite{bansal2018chauffeurnet,cui2019multimodal,djuric2020uncertainty,liang2020learning,gao2020vectornet,zhou2022hivt}. 
To the best of our knowledge, due to the data-hungry nature of deep learning, most of existing methods implicitly assume the availability of sufficiently rich and high-quality centralized datasets \cite{zhou2022hivt,liang2020learning,ngiam2021scene,gu2021densetnt,liu2021multimodal,weng2022whose,zhang2022adversarial,bahari2022vehicle}.  
This requirement easily leads to privacy leakage because raw trajectory data contains sensitive information such as personal information (home and company addresses, and driving logs) and vehicle information (vehicle types, brands, and appearances) \cite{glancy2012privacy}. The privacy threat quickly deteriorates as the need to acquire useful information across cities and/or countries \cite{didi2022} increases.  

Uncertainty-awareness becomes increasingly important for safety-crucial cyber physical systems like autonomous vehicles whose prediction module heavily relies on machine learning tools   \cite{zhang2016understanding,chiu2021probabilistic,su2022uncertainty,djuric2020uncertainty,tang2021collaborative,li2022uqnet,tang2022prediction}.
In general, there are two principal types of uncertainties. Data uncertainty (aleatoric) describes the intrinsically irreducible data variability. Model uncertainty (epistemic) refers to the shortcoming of the observational models that can be used to learn the underlying true mechanisms. In trajectory prediction literature, aleatoric uncertainty is often approximated by the variance estimate contained in the model output 
\cite{deo2018convolutional,djuric2020uncertainty,zhou2022hivt,tang2021collaborative}, while epistemic uncertainty is often modeled by Monte Carlo (MC) dropout \cite{gal2016dropout,djuric2020uncertainty} and ensembles \cite{lakshminarayanan2017simple,tang2022prediction}. 
Nevertheless, the potential of utilizing the intermediate uncertainty quantification in further improving trajectory prediction training is largely overlooked.

Federated Learning (FL) is a rapidly developing privacy-preserving decentralized learning framework in which a parameter server (PS) and a collection of clients collaboratively train a common model \cite{mcmahan2017communication,kairouz2021advances}.
In FL, instead of uploading data to the PS, the clients perform updates based on their own local data and periodically report their local updates to the PS. The PS then effectively aggregates those updates to obtain a fine-grained model and broadcasts the fine-grained model to the clients for further model updates.
Contextualize FL in the connected autonomous vehicles (CAVs) applications, each autonomous vehicle is a client which collects local data of their driving scenarios and the parameter server can be viewed as a computing center. 

\vskip 0.5\baselineskip 
\noindent Our contributions can be summarized as follows: \\
(1) We relax the raw data collection requirement 
by tailoring FL to connected autonomous vehicles to collaboratively train HiVT \cite{zhou2022hivt} -- a light-weight transformer; we name the resulting algorithm as {\em Federated Learning based Trajectory Prediction (FLTP)}. 
To incorporate uncertainty quantification, following the literature, we adopt the popular negative log-likelihood (NLL) of Laplace mixture distribution as the regression loss with location and scale parameters, respectively, decode the predicted trajectories and the corresponding aleatoric uncertainty. 
To the best of our knowledge, we are the first to apply FL on CAVs for collaborative trajectory prediction.
\\
(2) It is widely observed that \cite{mcmahan2017communication,kairouz2021advances} (also validated in our preliminary experimental results) that partial client participation can speed up the convergence and improve the accuracy.  
To further boost the performance of FLTP, we introduce {\em ALFLTP} which uses novel active learning techniques to carefully select the participating clients per iteration. We respectively consider the negative log-likelihood (NLL) and aleatoric uncertainty (AU) as client selection metrics. To the best of our knowledge, we are the first to consider using aleatoric uncertainty as a metric for client selection. \\
(3) Experiments on Argoverse dataset show that FLTP significantly outperforms the model trained on local data. In addition, compared with FLTP, 
ALFLTP-AU converges faster in training regression loss and performs better in terms of NLL, minADE and MR in most rounds. It also has more stable round-wise performance than ALFLTP-NLL.

\section{Related Work}

\noindent \textbf{Trajectory Prediction for Autonomous Vehicles.} 
The pipeline of trajectory prediction typically consists of three sub-tasks: input representation, context aggregation, and output representation. Input representation is often created through either rasterization \cite{bansal2018chauffeurnet,cui2019multimodal,djuric2020uncertainty} or vectorization \cite{liang2020learning,gao2020vectornet,zhou2022hivt}. 
Context aggregation modules are used to capture object interactions in  traffic such as vehicle-to-vehicle, vehicle-to-lane, and vehicle-to-pedestrian interactions; popular techniques include social pooling \cite{deo2018convolutional,song2020pip}, attention mechanism \cite{ngiam2021scene,yuan2021agentformer,liu2021multimodal,zhou2022hivt} or Graph Neural Network (GNN) \cite{li2019grip,liang2020learning,jeon2020scale}. 
To make multi-modal prediction for future trajectories, output representation often relies on approaches such as regression based approaches \cite{deo2018convolutional,liang2020learning,zhou2022hivt} and proposal based approaches \cite{chai2019multipath,phan2020covernet,zhao2021tnt,liu2021multimodal}. 


\noindent \textbf{Federated Learning for Trajectory Prediction.} 
FedAvg is the first and the most widely implemented FL algorithm \cite{mcmahan2017communication,kairouz2021advances}.
Despite FL has broad prospects in distributed information processing, only a few existing works adopt FL to tasks that are relevant to trajectory prediction for autonomous vehicles. Flow-FL \cite{majcherczyk2021flow} studies trajectory prediction 
for connected robot teams. ATPFL \cite{wang2022atpfl} combines automated machine learning and FL to automatically design human trajectory prediction models. To the best of our knowledge, applying FL on CAVs for trajectory prediction is not yet explored.

\noindent \textbf{Active Client Selection in Federated Learning.} 
The idea behind active learning is to identify data samples that are more informative for model training. 
Inspired by active learning, a handful existing works design active client selection strategy and demonstrate the power of such biased client selection \cite{goetz2019active,cho2020client,li2022uncertainty} with faster convergence. 
Specifically, \cite{goetz2019active,cho2020client} take local loss as the active learning metric and give clients with a higher active learning metric the priority to be selected. \cite{li2022uncertainty} adopts Bayesian active learning and takes model uncertainty as the metric. However, no existing client selection strategies use aleatoric uncertainty as a metric.

\section{Methods}
\subsection{Problem Formulation}
\begin{figure*}[ht]
\centering
\begin{subfigure}{0.8\linewidth}
\centering
    \includegraphics[width=1\linewidth]{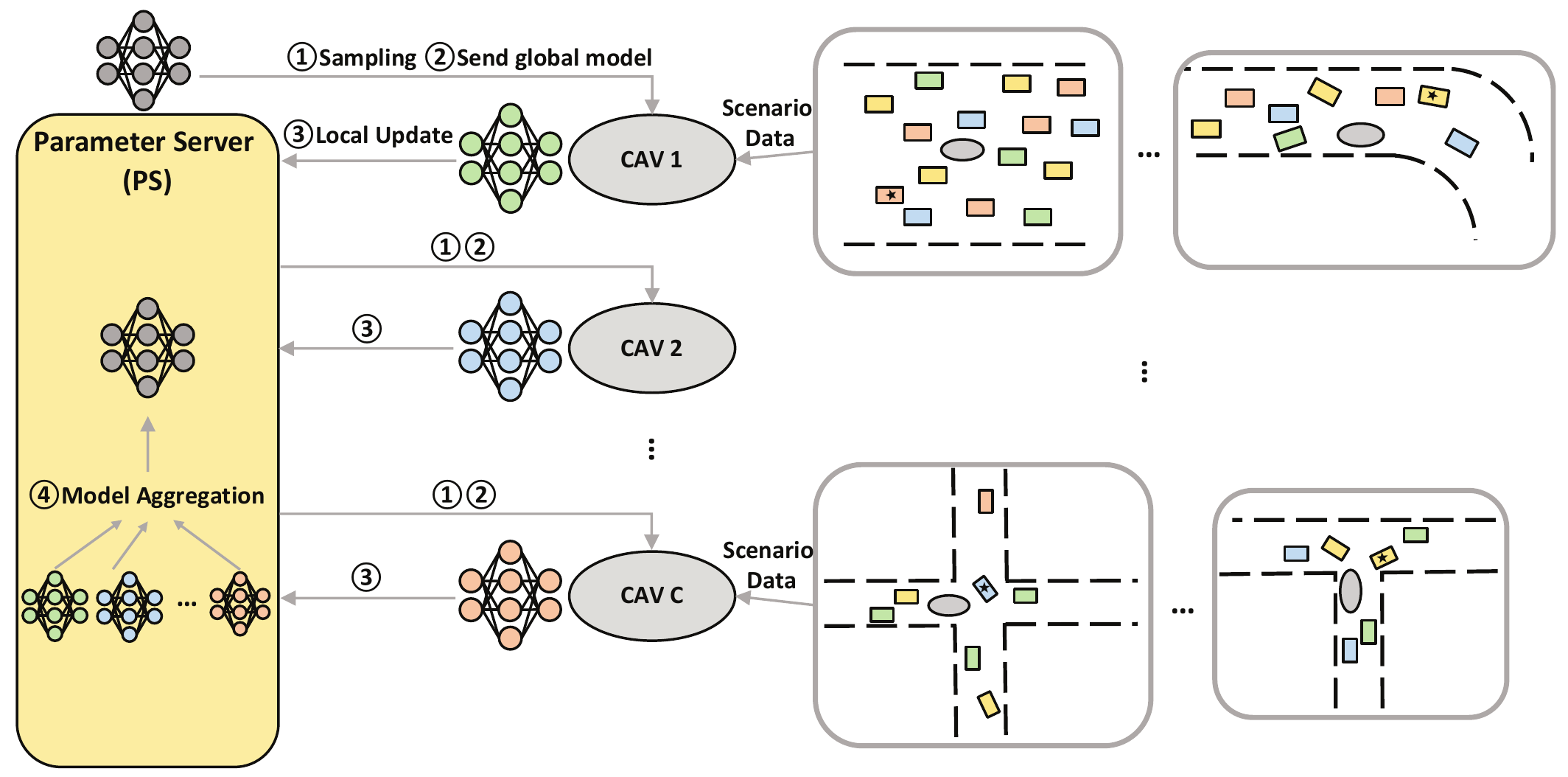}
\caption{The Framework of FLTP}
\label{fig:fltp}
\end{subfigure}

\begin{subfigure}{0.65\linewidth}
\centering
    \includegraphics[width=1\linewidth]{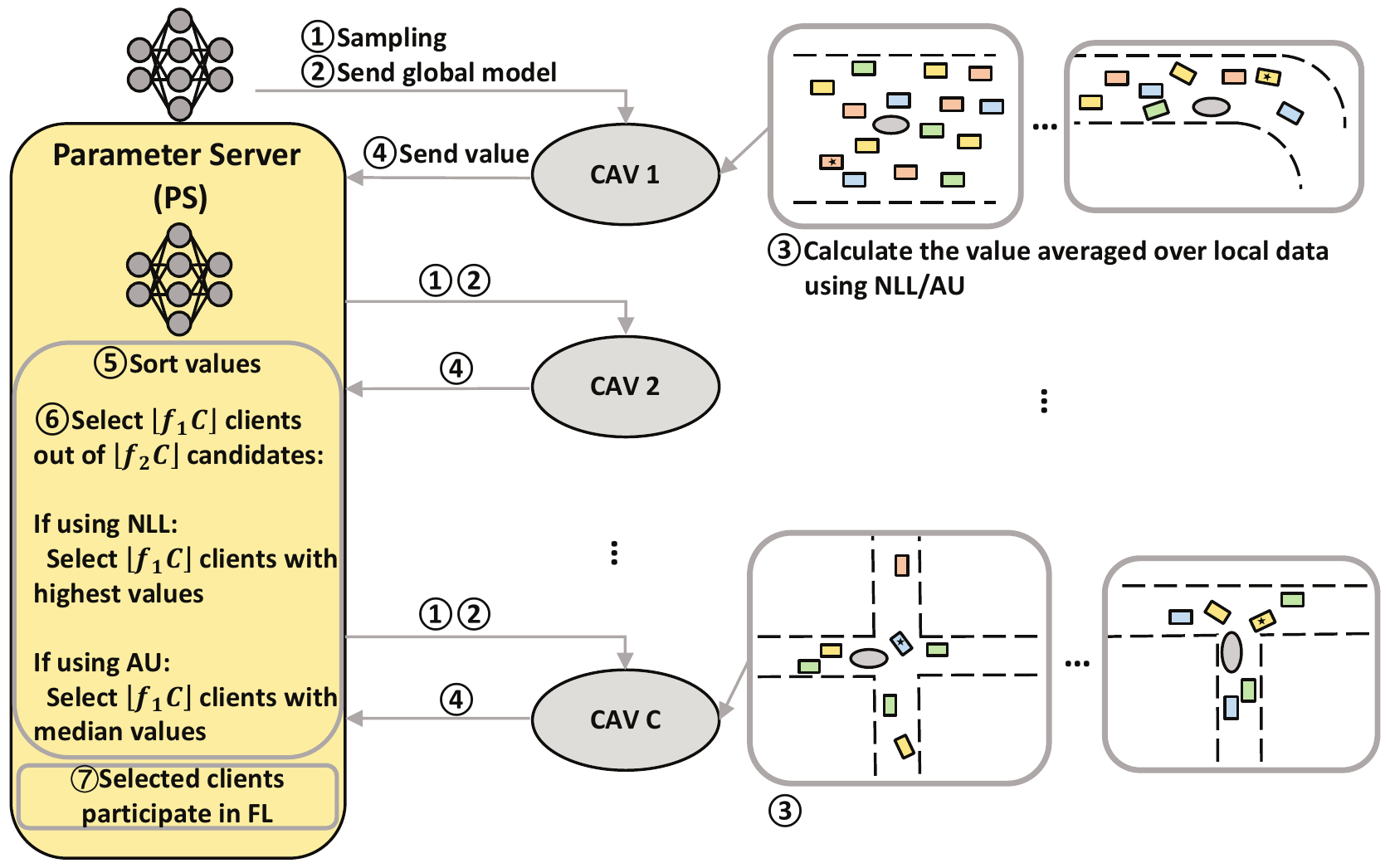}
\caption{The Framework of ALFLTP}
\label{fig:alfltp}
\end{subfigure}
\caption{Frameworks of FLTP and ALFLTP. Each gray oval represents a client (i.e., an ego vehicle) in FL. Each client collects its local driving scenario data using its sensors and cameras, does local updates, and communicates model weights with the central server. For each scenario,  the ego vehicle aims to predict the future trajectory of the target vehicle (starred) based on past trajectories of agents in the scenario and map information. In ALFLTP, clients are actively selected for computation of each round based on values of candidate clients measured by NLL or AU. Specifically, if we use NLL as the metric, clients with $m$ highest values are selected, while for using AU as the metric, clients with $m$ median values are selected.
}
\vspace{-10pt}
\end{figure*}

As shown in Fig.\ref{fig:fltp}, a driving scenario data (i.e.\,a data sample) 
can be described by a triple $S=(X, Y, \mathcal{M})$, where $X$ and $Y$ are the collections of observed and future trajectories of the involved agents (an agent can be a vehicle or a pedestrian), and $\mathcal{M}$ is the map information.  
Let $m$ denote the number of agents in the scenario, then $X$ and $Y$ can be expressed as $X = \{x_1,...,x_m$\} and $Y = \{y_1,...,y_m$\}, 
where $x_i \in \mathbb{R}^{2 \times T_{obs}}$ and $y_i \in \mathbb{R}^{2 \times T_{pre}}$ are the two-dimensional observed and future trajectory coordinates of agent $i$, with lengths $T_{obs}$ and $T_{pre}$, respectively. In particular, for any given scenario $S$, there is one target agent, denoted by $i^*$,  
among the $m$ agents which the ego vehicle is most interested in. In each scenario, the ego vehicle aims to predict the future trajectory of the target vehicle using $X$ and $\mathcal{M}$. 

The system contains a server and $C$ autonomous vehicles, each of which can collect its driving scenario data using Lidar sensors and cameras to  record the trajectories of all neighboring vehicles.   
We refer to each autonomous vehicle as one ego vehicle, which serves as one client in our FL framework. 
Each client $c\in \mathcal{C} \triangleq \{1, \cdots, C\}$ has a local dataset of size $K_c$, denoted as $\mathcal{D}_c = \{S^1, S^2,...,S^{K_c}\}$. Let $m_c^k$ be the number of agents in the $k$-th sample of client $c$. Let $K = \sum\limits_{c=1}^{C}K_c$ denote the total number of samples. 



\subsection{HiVT: Hierarchical Vector Transformer}
\label{subsec: HiVT}
Hierarchical Vector Transformer (HiVT) is a centralized, lightweight, and graph-based motion prediction model \cite{zhou2022hivt}. Towards scalability in the number of agents in the scene, HiVT decomposes the problem into local context extraction and global interaction modeling. HiVT achieves the state-of-the-art performance on the Argoverse motion forecasting benchmark. 
In this paper, we focus on training HiVT but under the FL framework. Different from the centralized HiVT, the local model updates at each client is done with respect to its local objective only. 

\vskip 0.5\baselineskip 
\noindent{\underline{\em Loss Function:}}  As HiVT makes prediction for all agents in a scenario in one single forward pass, prediction results of all agents will be used in the training loss. Nevertheless, in the inference stage only the prediction result of the target agent is evaluated. 
Similarly, in Section~\ref{subsec: active learning}, we do value calculation on the target agent only per driving scenario. 

HiVT uses the Laplace mixture probability density function as part of its loss function  
with $\hat{\mu}_{i,t,f}\in\mathbb{R}^2$ and $\hat{b}_{i,t,f}\in\mathbb{R}^2$, respectively, denoting the estimated location and scale parameters  for each agent $i\in\{1,...,m\}$ and each mixture component $f\in\{1,...,F\}$ at each prediction time step $t\in\{1,...,T_{pre}\}$. For any fixed $i$ and $t$, the two estimates $\hat{\mu}_{i,t,f}$ and $\hat{b}_{i,t,f}$ are interpreted as trajectory prediction and corresponding uncertainty, respectively, of the $f$-th predicted trajectory. 
The HiVT decoder also outputs predicted coefficients of the mixture model $\hat{P}_{i,f}\in \left[0,1\right]$ for each agent $i\in\{1,...,m\}$ and each mixture component $f\in\{1,...,F\}$.



HiVT only optimizes the best mode of $F$ trajectories. Specifically, the best trajectory for the $i$th agent is determined by the following equation:
\begin{equation}
      f_{best_i} = \underset{f\in\{1,...,F\}}{\arg\min}\sum\limits_{t=1}^{T_{pre}}\|y_{i,t}-\hat{\mu}_{i,t,f}\|_2. 
      \label{eq:f_best}
\end{equation}
The regression loss is the negative log-likelihood (NLL) of the Laplace distribution, which is shown as follows:
\begin{equation}
      L_{reg} = \frac{1}{m}\frac{1}{T_{pre}}\sum\limits_{i=1}^{m}\sum\limits_{t=1}^{T_{pre}}\left[\log(2\hat{b}_{i,t,f_{best_i}})+\frac{{\|y_{i,t}-\hat{\mu}_{i,t,f_{best_i}}\|}_1}{\hat{b}_{i,t,f_{best_i}}}\right]
      \label{eq:l_reg}
\end{equation}
The classification loss $L_{cls}$ is the cross entropy loss for optimizing mixture coefficients, which is shown as follows:
\begin{equation}
      L_{cls} = \frac{1}{m}\sum\limits_{i=1}^{m}\sum\limits_{f=1}^{F}-P_{i,f}\log\hat{P}_{i,f}
\end{equation}
with $$
   P_{i,f} = \frac{\exp(-\sum\limits_{t=1}^{T_{pre}}\|\hat{\mu}_{i,t,f}-y_{i,t}\|_2)}{\sum\limits_{j=1}^{F}{\exp(-\sum\limits_{t=1}^{T_{pre}}\|\hat{\mu}_{i,t,j}-y_{i,t}\|_2)}}
   \label{eq:soft_target}
$$
Then the final loss for scenario $S$ with model weight $w$ is:
\begin{equation}
    L(S, w) = L_{reg} + L_{cls}.
\end{equation}

\subsection{Federated Learning Based Trajectory Prediction (FLTP)}
\label{subsec: FLTP}
%
%
The loss function in Section \ref{subsec: HiVT} is defined for one driving scenario data.
In FL, as client can only get access to its local data, then the local objective is defined as:
\begin{equation}
\label{eq: local objective}
F_c(\mathcal{D}_c,w) = \frac{1}{K_c}\sum\limits_{S\in\mathcal{D}_c}L(S,w)
\end{equation}
 
We formally describe our FLTP in Algorithm \ref{alg:fltp_server}. 
It follows the general server-client interaction of FedAvg \cite{mcmahan2017communication}. Departing from the standard FedAvg, instead of stochastic gradient descent, we use AdamW as the local optimizer.

Specifically, in each global iteration:  
\begin{itemize}
\item The parameter server first randomly chooses $\lfloor f_1C \rfloor$ clients according to the probability vector $[K_1/K, \cdots, K_C/K]$ without replacement, where $f_1\in (0,1]$ is the client sampling rate given as algorithm input. For example, let $C=3$, $f_1 = \frac{2}{3}$, $K_1 = 1, K_2 = 2$, and $K_3=7$. The probability that client $1$ is chosen is $0.1+0.2\times \frac{0.1}{0.8}+0.7\times \frac{0.1}{0.3}$.
\item Then the parameter server sends the current model $w_r$ to each of the chosen client $\mathcal{L}_r$ to get further improvement on their local data. 
\item In parallel, each of the chosen client run AdamW with respect to Eq.\eqref{eq: local objective} with the specified minibatch size $B$ for $E$ epochs on local dataset. Concretely, in the \textsf{ClientUpdate} function, $\theta$ and $\Sigma$ are the weighted cumulative first and second moments, respectively, of the mini-batch gradients observed so far with $\beta_1, \beta_2 \in (0, 1)$ as the momentum parameters. Depending on the minibatch size $B$, for the first few iterations in the inner {\bf for}-loop, the smallest eigenvalues of $\hat{\Sigma}$ could be either zero or extermely small -- resulting in significant fluctuation of $w$. Hence, $\epsilon>0$ is used to smooth the updates. 
\item Finally, upon reception of the local updates $w_{r+1}^c$, the parameter servers aggregates those models accordingly to their relative local data volume to obtain $w_{r+1}$. 
\end{itemize}

It is worth noting that the algorithm can be improved via using global stepsize. We leave this direction to future work. 

\begin{algorithm}
        \caption{FLTP}
	\label{alg:fltp_server}
	\begin{algorithmic}[1]
	\renewcommand{\algorithmicrequire}{\textbf{Input:}}
        \REQUIRE 
        initial model $w_0$, number of clients $C$, client sampling rate $f_1\in (0, 1]$, local data volume $\{K_1, \cdots, K_C\}$, 
        stepsize $\eta$, batch size $B$, number of epoch $E$, weight decay $\lambda$,  momentum parameters $\beta_1, \beta_2\in (0,1)$, smooth parameter $\epsilon$;
        \renewcommand{\algorithmicensure}{\textbf{Output:}}
        \ENSURE $w_R$; 
        \STATE Initialization: $w \leftarrow w_0$; 
        \FOR{each round $r=1$ to $R$}
        \STATE Randomly sample $\lfloor f_1C \rfloor$ clients according to the probability vector $[K_1/K, \cdots, K_C/K]$ without replacement. Let $\mathcal{L}_r$ denote the resulting set of clients;  
        \STATE Send $w_{r-1}$ to each of the chosen client in $\mathcal{L}_r$; 
        \FOR {each client $c \in \mathcal{L}_r$ \textbf{in parallel}}
        \STATE $w_{r}^c \leftarrow$ \textsf{ClientUpdate}$(w_r, \eta, B, E, \beta_1, \beta_2, \epsilon)$; 
        \ENDFOR
        \STATE $w_{r} \leftarrow \sum\limits_{c\in \mathcal{L}_r} \frac{K_{c}}{\tilde{K}_{r}}w_{r}^c$, where $\tilde{K}_r \triangleq \sum\limits_{c\in \mathcal{L}_r} K_c$; 
        \ENDFOR
        \RETURN $w_R$;  
        \\\hrulefill
 \STATE \textsf{ClientUpdate}$(w, \eta, B, E, \beta_1, \beta_2, \epsilon)$
 \STATE Initialization: first moment $\theta \leftarrow 0$, second moment $\Sigma \leftarrow 0$, counter $t \leftarrow 0$; 
 \FOR {each local epoch $i=1, \cdots, E$}
        \STATE $\mathcal{B}_c \leftarrow $ divide $\mathcal{D}_c$ into batches with batch size $B$
        \FOR{each batch $b \in \mathcal{B}_c$}
        \STATE $t \leftarrow t + 1$
        \STATE $g = \nabla F_c(b,w)$
        \STATE $w \leftarrow w - \eta\lambda g$
        \STATE $\theta \leftarrow \beta_1\theta + (1-\beta_1)g$, $\Sigma \leftarrow \beta_2\Sigma + (1-\beta_2)gg^{\top}$
        \STATE $\hat{\theta} \leftarrow \frac{\theta}{1-\beta_1^t}$, $\hat{\Sigma} \leftarrow \frac{\Sigma}{1-\beta_2^t}$
        \STATE   $w \leftarrow w - \eta \left(\hat{\Sigma}^{1/2} + \epsilon I\right)^{-1} \hat{\theta}$   
        \ENDFOR
        \ENDFOR
        \RETURN $w$
	\end{algorithmic}
\end{algorithm}
\setlength{\floatsep}{0.1cm}

\subsection{Active Learning-boosted FLTP (ALFLTP)}
\label{subsec: active learning}
We use active learning to carefully select the clients to participate in each iteration. 
Departing from existing literature \cite{cho2020client}, instead of directly using the whole loss function as the metric, we consider two types of uncertainty-aware client selection metrics: negative log-likelihood (NLL) and aleatoric uncertainty (AU). 

In our Algorithm \ref{alg:alfltp}, each client has a value variable $v_i$. In each round $r$ (where $r>2$), the parameter server samples the clients twice.  It first randomly chooses $\lfloor f_2C \rfloor$ clients as it does in Algorithm \ref{alg:fltp_server}. Each of the chosen clients in $\mathcal{Q}_r$ updates its local value $v$ according to 
$$
G_c(\mathcal{D}_c,w) = \frac{1}{K_c}\sum\limits_{S\in\mathcal{D}_c}G(S,w)
$$
where 
\begin{align*}
G(S,w)= 
 \begin{cases}
 \frac{1}{T_{pre}}\sum\limits_{t=1}^{T_{pre}}&\left[\log(2\hat{b}_{i^*,t,f_{best_{i^*}}})+\frac{{\|y_{i,t}-\hat{\mu}_{i^*,t,f_{best_{i^*}}}\|}_1} {\hat{b}_{i^*,t,f_{best_{i^*}}}} \right] \qquad \qquad \text{if }NLL \\
\frac{1}{T_{pre}}\sum\limits_{t=1}^{T_{pre}}&\hat{b}_{i^*,t,f_{best_{i^*}}}\qquad \qquad \text{if }AU\\ 
 \end{cases}
\end{align*}
where $\hat{\mu}_{i^*,t,f_{best_{i^*}}}$ denotes prediction location of the target agent at time step $t$ from the best mode of $F$ trajectories and $\hat{b}_{i^*,t,f_{best_{i^*}}}$ denotes the corresponding aleatoric uncertainty.
All clients that are not contained in $\mathcal{Q}_r$ reset their values $v_i=0$.  
If NLL is used as the selection metric, then the parameter server chooses the set $\mathcal{L}_r$ to be the $\lfloor f_1C \rfloor$ clients with highest $v_i$; If AU is used as the metric, then the parameter server chooses $\mathcal{L}_r$ to contains the $\lfloor f_1C \rfloor$ whose values $v_i$ are closest to their median. 
When $r=1$ (i.e., lines 2-6 in Algorithm \ref{alg:alfltp}), as there is no global model from the last round for value calculation, client selection method is similar to that in FLTP.

\begin{algorithm}
        \caption{ALFLTP}
        \label{alg:alfltp}
	\begin{algorithmic}[1]
	\renewcommand{\algorithmicrequire}{\textbf{Input:}}
        \REQUIRE  initial model $w_0$, stepsize $\eta$, number of epoch $E$, client sampling rate $f_1\in (0, 1]$, candidate sampling rate $f_2\in (0, 1]$, local data volume $\{K_1, \cdots, K_C\}$, 
         weight decay $\lambda$,  momentum parameters $\beta_1, \beta_2\in (0,1)$, smooth parameter $\epsilon$;
        \renewcommand{\algorithmicensure}{\textbf{Output:}}
        \ENSURE $w_R$
        \STATE Initialization $w \leftarrow w_0$
        \STATE The PS randomly samples $\lfloor f_1C \rfloor$ clients according to the probability vector $[K_1/K, \cdots, K_C/K]$ without replacement. Let $\mathcal{L}_r$ denote the resulting set of clients;  
        \FOR {each client $c \in {\mathcal{L}}_1$ \textbf{in parallel}}
        \STATE $w_{1}^c \leftarrow$ ClientUpdate$(w, \eta, B, E, \beta_1, \beta_2, \epsilon)$ 
        \ENDFOR
        \STATE $w_{1} \leftarrow \sum\limits_{c\in \mathcal{L}_1} \frac{K_{c}}{\tilde{K}_1}w_{1}^c$, where $\tilde{K}_1 \triangleq \sum\limits_{c\in \mathcal{L}_1} K_c$
        \FOR{each round $r=2$ to $R$}
        \STATE The PS selects $\lfloor f_2C \rfloor$ clients randomly as in line 2. Let $\mathcal{Q}_r$ denote the resulting client set; 
        \STATE The PS broadcasts $w_{r-1}$ to each client in $\mathcal{Q}_r$; 
        \FOR {each client $c$ \textbf{in parallel}}
        \IF{$c\in \mathcal{Q}_r$}
        \STATE $v_c \leftarrow G_c({\mathcal{D}_{c}},w_{r-1})$; 
        \ELSE 
        \STATE $v_c \leftarrow 0$
        \ENDIF
        
        Reports $v_c$ to the PS; 
        \ENDFOR
        \STATE The PS sorts $\{v_c\}_{c\in \mathcal{C}}$\;  
        \IF{u = NLL}
        \STATE The PS selects $\lfloor f_1C \rfloor$ clients with highest values $v_i$, denoting the resulting set as $\mathcal{L}_{r}$; 
        \ELSE  
        \STATE The PS selects $\lfloor f_1C \rfloor$ clients with values that closest to the median values $v_i$, denoting the resulting set as $\mathcal{L}_{r}$; 
        \ENDIF
        \FOR {each client $c \in {\mathcal{L}}_r$ \textbf{in parallel}}
        \STATE $w_{r}^c \leftarrow$ ClientUpdate$(w, \eta, B, E, \beta_1, \beta_2, \epsilon)$ 
        \ENDFOR
        \STATE $w_{r} \leftarrow \sum\limits_{c\in \mathcal{L}_r} \frac{K_{c}}{\tilde{K}_r}w_{r-1}^c$, where $\tilde{K}_r \triangleq \sum\limits_{c\in \mathcal{L}_r} K_c$.
        \ENDFOR
	\end{algorithmic}
\end{algorithm}
\setlength{\floatsep}{0.1cm}

\paragraph{NLL as a selection metric}
We choose NLL as one selction metric for the following two reasons: 
Since NLL is incorporated as part of the loss function, a client has a higher NLL if the global model is not sufficiently trained with respect to its local data.
As the local data is non-iid covering different driving scenarios, by selecting clients with higher NLL, the global model in FL is encouraged to do more local training on clients with more difficult data. 

\paragraph{AU as a selection metric} 
This metric is inspired by \cite{xi2021robust}, where incremental active learning is adopted for human trajectory prediction to evaluate candidate data samples and then select more valuable samples. Specifically, both noisy and redundant trajectory candidate data samples are removed and the model trained on filtered data samples achieves better performance. In our ALFLTP, we exploit aleatoric uncertainty to measure the degree of data noise. High aleatoric uncertainty means data are very noisy, while low aleatoric uncertainty means data are easy and the model is certain about them. As a result, we prefer clients with median aleatoric uncertainty, as data on these clients are both representative and less noisy. 

\vskip 0.6\baselineskip
\noindent {\em Relaxing full client participation in updating $v$.} 
For ease of exposition, in lines 10-15 of Algorithm \ref{alg:alfltp}, we let every client participate in updating $v_c$. In practice, it suffices to have the clients in $\mathcal{Q}_r$ do the value updates only. Since the value update does not rely on any previous value of $v_c$,
the updated values are only used in the sorting at the PS, and the PS knows the $\mathcal{Q}_r$, the PS can treat $v_c=0$ for all $c\notin \mathcal{Q}_r$.

\section{Experiments}
\subsection{Experimental Setup}
\noindent \textbf{Dataset:} We use Argoverse Motion Forecasting v1.1 dataset for training and evaluation. In order to simulate distributed trajectory data for federated learning, we distribute the training set to 100 clients based on the city label of each data sample (driving scenario), where 95521 samples are from Pittsburgh and 110421 samples are from Miami. More specifically, the samples from the two cites, are evenly distributed to 50 clients, 
with each client denoting an autonomous vehicle that can collect and process traffic data in its area. The validation set contains 39472 samples. All training and validation scenarios consist of trajectories of 5 seconds sampled at 10 Hz and map information. The Argoverse Motion Forecasting challenge is to predict future trajectories of 3 seconds of focal agents with past trajectories of 2 seconds as inputs.

\noindent \textbf{Model and Training Parameters:} We use HiVT \cite{zhou2022hivt} with 64 hidden dimensions as our trajectory prediction model, which is a light-weight transformer based model. We use similar parameter settings for local HiVT models in FLTP and ALFLTP as the centralized HiVT. Specifically, for each local model in FLTP, learning rate $\eta$, weight decay, dropout rate, local batchsize $B$, local epochs $E$ and local optimizer are set to be $5 \times 10^{-4}$, $1 \times 10^{-4}$, $0.1$, $32$, $4$ and AdamW. We train FLTP and ALFLTP for 250 rounds. Fraction of clients selected for communication in each round $f_1$ is set to be 0.1.

\noindent \textbf{Evaluation Metrics:} We use NLL, Minimum Average Displacement Error (minADE), Minimum Final Displacement Error (minFDE) and Miss Rate (MR) to evaluate model performance quantitatively. minADE measures the average L2 distance between the best predicted trajectory (the trajectory with the minimum error at the endpoint) and the ground truth. minFDE measures the endpoint L2 distance between the best predicted trajectory and the ground truth. MR measures the fraction of the number of scenarios where endpoint errors of all predicted trajectories are larger than 2 meters.

\subsection{FLTP v.s. training on local data}
We quantitatively compare the global model of FLTP and the local model of an arbitrarily chosen client  when it does not participate in communication and only updates using its local data. Here we have chosen the client 0; selecting any other client would yield the same result. As is shown in Fig. \ref{fig:wofl} and Table \ref{i^*b:roundshotdata}, FLTP significantly outperforms the client without FL, demonstrating the effectiveness of FLTP exploiting multi-source traffic data though it does not explicitly access raw local data. Specifically, after about 50 rounds, the local model of client 0 begins to exhibit worse performance as the number of training rounds increases, indicating the local model without FL has poor generalization.

\begin{figure*}[ht]
\centering
\begin{subfigure}{0.27\linewidth}
\begin{tikzpicture}
\begin{axis}[
    xlabel={\scriptsize{Round}},
    ylabel={\scriptsize{$L_{reg}$}},
    every axis x label/.style={at={(current axis.south)},below=8pt},
    every axis y label/.style={
            at={(ticklabel* cs:0.5)},rotate=90,anchor=center,align=center, above=10pt},
    xmin=0, xmax=250,
    ymin=-0.5, ymax=0,
    xtick={0,50,100,150,200,250},
    xticklabels={\scriptsize{0},\scriptsize{50},\scriptsize{100},\scriptsize{150},\scriptsize{200},\scriptsize{250}},
    ytick={-0.5,-0.4,-0.3,-0.2,-0.1,0},
    yticklabels={\scriptsize{-0.5},\scriptsize{-0.4},\scriptsize{-0.3},\scriptsize{-0.2},\scriptsize{-0.1},\scriptsize{0}},
    height = 1\linewidth,
    width = 1\linewidth,
    xmajorgrids=true,
    ymajorgrids=true,
    grid style=dashed,
    legend cell align={left},
    legend style={inner sep=0pt,row sep=-3pt},
    legend style={font=\tiny},
    legend style={nodes={scale=0.7, transform shape}},
    every axis plot/.append style={ultra thick}
]
    
    \addplot [
    color=Gray,
    line width=1pt
    ]
    table[x=roundshow,y=flloss,col sep = comma] {iros_data_noniid.csv};

    \addplot [
    color=Salmon,
    line width=1pt,
    ]
    table[x=roundshow,y=localloss,col sep = comma] {iros_data_noniid.csv};

\legend{FLTP, Client 0 w/o FL}

\end{axis}
\end{tikzpicture}
\vspace{-5pt}
\end{subfigure}
\hspace{-20pt}
\begin{subfigure}{0.27\linewidth}
\begin{tikzpicture}
\begin{axis}[
    xlabel={\scriptsize{Round}},
    ylabel={\scriptsize{minADE}},
    every axis x label/.style={at={(current axis.south)},below=8pt},
    every axis y label/.style={
            at={(ticklabel* cs:0.5)},rotate=90,anchor=center,align=center, above=13pt},
    xmin=0, xmax=250,
    ymin=0.7, ymax=1.5,
    xtick={0,50,100,150,200,250},
    xticklabels={\scriptsize{0},\scriptsize{50},\scriptsize{100},\scriptsize{150},\scriptsize{200},\scriptsize{250}},
    ytick={0.7,0.9,1.1,1.3,1.5},
    yticklabels={\scriptsize{0.7},\scriptsize{0.9},\scriptsize{1.1},\scriptsize{1.3},\scriptsize{1.5}},
    height = 1\linewidth,
    xmajorgrids=true,
    ymajorgrids=true,
    grid style=dashed,
    legend cell align={left},
    legend style={inner sep=0pt,row sep=-3pt},
    legend style={font=\tiny},
    legend style={nodes={scale=0.7, transform shape}},
    every axis plot/.append style={ultra thick}
]
    
    \addplot [
    color=Gray,
    line width=1pt,
    ]
    table[x=roundshow,y=flade,col sep = comma] {iros_data_noniid.csv};

    \addplot [
    color=Salmon,
    line width=1pt,
    ]
    table[x=roundshow,y=localade,col sep = comma] {iros_data_noniid.csv};

\legend{FLTP, Client 0 w/o FL}
    
\end{axis}
\end{tikzpicture}
\vspace{-5pt}
\end{subfigure}
\hspace{-20pt}
\begin{subfigure}{0.27\linewidth}
\begin{tikzpicture}
\begin{axis}[
    xlabel={\scriptsize{Round}},
    ylabel={\scriptsize{minFDE}},
    every axis x label/.style={at={(current axis.south)},below=8pt},
    every axis y label/.style={
            at={(ticklabel* cs:0.5)},rotate=90,anchor=center,align=center, above=10pt},
    xmin=0, xmax=250,
    ymin=1.1, ymax=2.3,
    xtick={0,50,100,150,200,250},
    xticklabels={\scriptsize{0},\scriptsize{50},\scriptsize{100},\scriptsize{150},\scriptsize{200},\scriptsize{250}},
    ytick={1.1,1.3,1.5,1.7,1.9,2.1,2.3},
    yticklabels={\scriptsize{1.1},\scriptsize{1.3},\scriptsize{1.5},\scriptsize{1.7},\scriptsize{1.9},\scriptsize{2.1},\scriptsize{2.3}},
    height = 1\linewidth,
    xmajorgrids=true,
    ymajorgrids=true,
    grid style=dashed,
    legend cell align={left},
    legend style={inner sep=0pt,row sep=-3pt},
    legend style={font=\tiny},
    legend style={nodes={scale=0.7, transform shape}},
    every axis plot/.append style={ultra thick}
]
    
    \addplot [
    color=Gray,
    line width=1pt,
    ]
    table[x=roundshow,y=flfde,col sep = comma] {iros_data_noniid.csv};

    \addplot [
    color=Salmon,
    line width=1pt,
    ]
    table[x=roundshow,y=localfde,col sep = comma] {iros_data_noniid.csv};

\legend{FLTP, Client 0 w/o FL}
    
\end{axis}
\end{tikzpicture}
\vspace{-5pt}
\end{subfigure}
\hspace{-20pt}
\begin{subfigure}{0.27\linewidth}
\begin{tikzpicture}
\begin{axis}[
    xlabel={\scriptsize{Round}},
    ylabel={\scriptsize{MR}},
    every axis x label/.style={at={(current axis.south)},below=8pt},
    every axis y label/.style={
            at={(ticklabel* cs:0.5)},rotate=90,anchor=center,align=center, above=13pt},
    xmin=0, xmax=250,
    ymin=0.1, ymax=0.5,
    xtick={0,50,100,150,200,250},
    xticklabels={\scriptsize{0},\scriptsize{50},\scriptsize{100},\scriptsize{150},\scriptsize{200},\scriptsize{250}},
    ytick={0.1,0.2,0.3,0.4,0.5},
    yticklabels={\scriptsize{0.1},\scriptsize{0.2},\scriptsize{0.3},\scriptsize{0.4},\scriptsize{0.5}},
    height = 1\linewidth,
    xmajorgrids=true,
    ymajorgrids=true,
    grid style=dashed,
    legend cell align={left},
    legend style={inner sep=0pt,row sep=-3pt},
    legend style={font=\tiny},
    legend style={nodes={scale=0.7, transform shape}},
    every axis plot/.append style={ultra thick}
]
    
    \addplot [
    color=Gray,
    line width=1pt,
    ]
    table[x=roundshow,y=flmr,col sep = comma] {iros_data_noniid.csv};

    \addplot [
    color=Salmon,
    line width=1pt,
    ]
    table[x=roundshow,y=localmr,col sep = comma] {iros_data_noniid.csv};

\legend{FLTP, Client 0 w/o FL}
    
\end{axis}
\end{tikzpicture}
\vspace{-5pt}
\end{subfigure}
\caption{Round-wise comparison between FLTP and the local model of client 0 without FL. Fraction of clients selected for communication in each round $f_1$ is set to be 0.1.}
\label{fig:wofl}
\vspace{10pt}
\end{figure*}
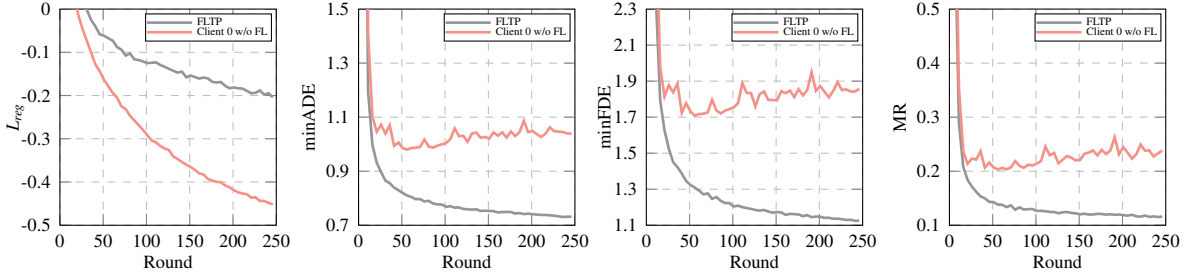

\begin{figure*}[ht]
\centering
\begin{subfigure}{0.27\linewidth}
\begin{tikzpicture}
\begin{axis}[
    xlabel={\scriptsize{Round}},
    ylabel={\scriptsize{$L_{reg}$}},
    every axis x label/.style={at={(current axis.south)},below=8pt},
    every axis y label/.style={
            at={(ticklabel* cs:0.5)},rotate=90,anchor=center,align=center, above=10pt},
    xmin=0, xmax=250,
    ymin=-0.22, ymax=0,
    xtick={0,50,100,150,200,250},
    xticklabels={\scriptsize{0},\scriptsize{50},\scriptsize{100},\scriptsize{150},\scriptsize{200},\scriptsize{250}},
    ytick={-0.2,-0.1,0,0.1,0.2,0.3,0.4,0.5,0.6},
    yticklabels={\scriptsize{-0.2},\scriptsize{-0.1},\scriptsize{0},\scriptsize{0.1},\scriptsize{0.2},\scriptsize{0.3},\scriptsize{0.3},\scriptsize{0.4},\scriptsize{0.5},\scriptsize{0.6}},
    height = 1\linewidth,
    width = 1\linewidth,
    xmajorgrids=true,
    ymajorgrids=true,
    grid style=dashed,
    legend cell align={left},
    legend style={inner sep=0pt,row sep=-3pt},
    legend style={font=\tiny},
    legend style={nodes={scale=0.7, transform shape}},
    every axis plot/.append style={ultra thick}
]
    
    \addplot [
    color=Gray,
    line width=1pt
    ]
    table[x=roundshow,y=flloss,col sep = comma] {iros_data_noniid.csv};

    \addplot [
    color=BurntOrange,
    line width=1pt,
    ]
    table[x=roundshow,y=flalloss0.15,col sep = comma] {iros_data_noniid.csv};

    \addplot [
    color=Goldenrod,
    line width=1pt,
    ]
    table[x=roundshow,y=flalloss0.3,col sep = comma] {iros_data_noniid.csv};

\legend{FLTP,ALFLTP-NLL($f_2$=0.15),ALFLTP-NLL($f_2$=0.30)}

\end{axis}
\end{tikzpicture}
\vspace{-5pt}
\end{subfigure}
\hspace{-20pt}
\begin{subfigure}{0.27\linewidth}
\begin{tikzpicture}
\begin{axis}[
    xlabel={\scriptsize{Round}},
    ylabel={\scriptsize{minADE}},
    every axis x label/.style={at={(current axis.south)},below=8pt},
    every axis y label/.style={
            at={(ticklabel* cs:0.5)},rotate=90,anchor=center,align=center, above=13pt},
    xmin=0, xmax=250,
    ymin=0.72, ymax=0.85,
    xtick={0,50,100,150,200,250},
    xticklabels={\scriptsize{0},\scriptsize{50},\scriptsize{100},\scriptsize{150},\scriptsize{200},\scriptsize{250}},
    ytick={0.72,0.74,0.76,0.78,0.80,0.82,0.84},
    yticklabels={\scriptsize{0.72},\scriptsize{0.74},\scriptsize{0.76},\scriptsize{0.78},\scriptsize{0.80},\scriptsize{0.82},\scriptsize{0.84}},
    height = 1\linewidth,
    xmajorgrids=true,
    ymajorgrids=true,
    grid style=dashed,
    legend cell align={left},
    legend style={inner sep=0pt,row sep=-3pt},
    legend style={font=\tiny},
    legend style={nodes={scale=0.7, transform shape}},
    every axis plot/.append style={ultra thick}
]
    
    \addplot [
    color=Gray,
    line width=1pt,
    ]
    table[x=roundshow,y=flade,col sep = comma] {iros_data_noniid.csv};

    \addplot [
    color=BurntOrange,
    line width=1pt,
    ]
    table[x=roundshow,y=flalade0.15,col sep = comma] {iros_data_noniid.csv};

    \addplot [
    color=Goldenrod,
    line width=1pt,
    ]
    table[x=roundshow,y=flalade0.3,col sep = comma] {iros_data_noniid.csv};



\legend{FLTP,ALFLTP-NLL($f_2$=0.15),ALFLTP-NLL($f_2$=0.30)}
    
\end{axis}
\end{tikzpicture}
\vspace{-5pt}
\end{subfigure}
\hspace{-20pt}
\begin{subfigure}{0.27\linewidth}
\begin{tikzpicture}
\begin{axis}[
    xlabel={\scriptsize{Round}},
    ylabel={\scriptsize{minFDE}},
    every axis x label/.style={at={(current axis.south)},below=8pt},
    every axis y label/.style={
            at={(ticklabel* cs:0.5)},rotate=90,anchor=center,align=center, above=10pt},
    xmin=0, xmax=250,
    ymin=1.1, ymax=1.4,
    xtick={0,50,100,150,200,250},
    xticklabels={\scriptsize{0},\scriptsize{50},\scriptsize{100},\scriptsize{150},\scriptsize{200},\scriptsize{250}},
    ytick={1.1,1.2,1.3,1.4},
    yticklabels={\scriptsize{1.1},\scriptsize{1.2},\scriptsize{1.3},\scriptsize{1.4}},
    height = 1\linewidth,
    xmajorgrids=true,
    ymajorgrids=true,
    grid style=dashed,
    legend cell align={left},
    legend style={inner sep=0pt,row sep=-3pt},
    legend style={font=\tiny},
    legend style={nodes={scale=0.7, transform shape}},
    every axis plot/.append style={ultra thick}
]
    
    \addplot [
    color=Gray,
    line width=1pt,
    ]
    table[x=roundshow,y=flfde,col sep = comma] {iros_data_noniid.csv};

    \addplot [
    color=BurntOrange,
    line width=1pt,
    ]
    table[x=roundshow,y=flalfde0.15,col sep = comma] {iros_data_noniid.csv};

    \addplot [
    color=Goldenrod,
    line width=1pt,
    ]
    table[x=roundshow,y=flalfde0.3,col sep = comma] {iros_data_noniid.csv};

    

\legend{FLTP,ALFLTP-NLL($f_2$=0.15),ALFLTP-NLL($f_2$=0.30)}
    
\end{axis}
\end{tikzpicture}
\vspace{-5pt}
\end{subfigure}
\hspace{-20pt}
\begin{subfigure}{0.27\linewidth}
\begin{tikzpicture}
\begin{axis}[
    xlabel={\scriptsize{Round}},
    ylabel={\scriptsize{MR}},
    every axis x label/.style={at={(current axis.south)},below=8pt},
    every axis y label/.style={
            at={(ticklabel* cs:0.5)},rotate=90,anchor=center,align=center, above=13pt},
    xmin=0, xmax=250,
    ymin=0.11, ymax=0.15,
    xtick={0,50,100,150,200,250},
    xticklabels={\scriptsize{0},\scriptsize{50},\scriptsize{100},\scriptsize{150},\scriptsize{200},\scriptsize{250}},
    ytick={0.11,0.12,0.13,0.14,0.15},
    yticklabels={\scriptsize{0.11},\scriptsize{0.12},\scriptsize{0.13},\scriptsize{0.14},\scriptsize{0.15}},
    height = 1\linewidth,
    xmajorgrids=true,
    ymajorgrids=true,
    grid style=dashed,
    legend cell align={left},
    legend style={inner sep=0pt,row sep=-3pt},
    legend style={font=\tiny},
    legend style={nodes={scale=0.7, transform shape}},
    every axis plot/.append style={ultra thick}
]
    
    \addplot [
    color=Gray,
    line width=1pt,
    ]
    table[x=roundshow,y=flmr,col sep = comma] {iros_data_noniid.csv};

    \addplot [
    color=BurntOrange,
    line width=1pt,
    ]
    table[x=roundshow,y=flalmr0.15,col sep = comma] {iros_data_noniid.csv};

    \addplot [
    color=Goldenrod,
    line width=1pt,
    ]
    table[x=roundshow,y=flalmr0.3,col sep = comma] {iros_data_noniid.csv};



\legend{FLTP,ALFLTP-NLL($f_2$=0.15),ALFLTP-NLL($f_2$=0.30)}
    
\end{axis}
\end{tikzpicture}
\vspace{-5pt}
\end{subfigure}
\caption{Round-wise comparison between FLTP and ALFLTP-NLL}
\label{fig:alfltpnll}
\end{figure*}
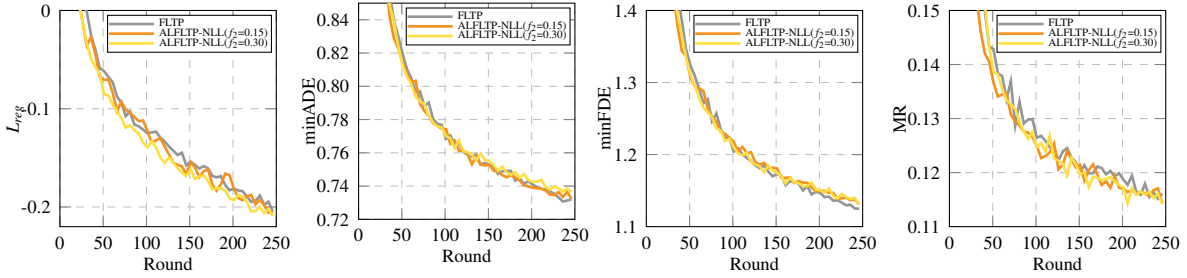

\begin{figure*}[ht]
\centering
\begin{subfigure}{0.27\linewidth}
\begin{tikzpicture}
\begin{axis}[
    xlabel={\scriptsize{Round}},
    ylabel={\scriptsize{$L_{reg}$}},
    every axis x label/.style={at={(current axis.south)},below=8pt},
    every axis y label/.style={
            at={(ticklabel* cs:0.5)},rotate=90,anchor=center,align=center, above=10pt},
    xmin=0, xmax=250,
    ymin=-0.22, ymax=0,
    xtick={0,50,100,150,200,250},
    xticklabels={\scriptsize{0},\scriptsize{50},\scriptsize{100},\scriptsize{150},\scriptsize{200},\scriptsize{250}},
    ytick={-0.2,-0.1,0,0.1,0.2,0.3,0.4,0.5,0.6},
    yticklabels={\scriptsize{-0.2},\scriptsize{-0.1},\scriptsize{0},\scriptsize{0.1},\scriptsize{0.2},\scriptsize{0.3},\scriptsize{0.3},\scriptsize{0.4},\scriptsize{0.5},\scriptsize{0.6}},
    height = 1\linewidth,
    width = 1\linewidth,
    xmajorgrids=true,
    ymajorgrids=true,
    grid style=dashed,
    legend cell align={left},
    legend style={inner sep=0pt,row sep=-3pt},
    legend style={font=\tiny},
    legend style={nodes={scale=0.7, transform shape}},
    every axis plot/.append style={ultra thick}
]
    
    \addplot [
    color=Gray,
    line width=1pt
    ]
    table[x=roundshow,y=flloss,col sep = comma] {iros_data_noniid.csv};



    \addplot [
    color=LimeGreen,
    line width=1pt,
    ]
    table[x=roundshow,y=auloss0.15,col sep = comma] {iros_data_noniid.csv};

    \addplot [
    color=Cyan,
    line width=1pt,
    ]
    table[x=roundshow,y=auloss0.3,col sep = comma] {iros_data_noniid.csv};

\legend{FLTP,ALFLTP-AU($f_2$=0.15),ALFLTP-AU($f_2$=0.30)}

\end{axis}
\end{tikzpicture}
\vspace{-5pt}
\end{subfigure}
\hspace{-20pt}
\begin{subfigure}{0.27\linewidth}
\begin{tikzpicture}
\begin{axis}[
    xlabel={\scriptsize{Round}},
    ylabel={\scriptsize{minADE}},
    every axis x label/.style={at={(current axis.south)},below=8pt},
    every axis y label/.style={
            at={(ticklabel* cs:0.5)},rotate=90,anchor=center,align=center, above=13pt},
    xmin=0, xmax=250,
    ymin=0.72, ymax=0.85,
    xtick={0,50,100,150,200,250},
    xticklabels={\scriptsize{0},\scriptsize{50},\scriptsize{100},\scriptsize{150},\scriptsize{200},\scriptsize{250}},
    ytick={0.72,0.74,0.76,0.78,0.80,0.82,0.84},
    yticklabels={\scriptsize{0.72},\scriptsize{0.74},\scriptsize{0.76},\scriptsize{0.78},\scriptsize{0.80},\scriptsize{0.82},\scriptsize{0.84}},
    height = 1\linewidth,
    xmajorgrids=true,
    ymajorgrids=true,
    grid style=dashed,
    legend cell align={left},
    legend style={inner sep=0pt,row sep=-3pt},
    legend style={font=\tiny},
    legend style={nodes={scale=0.7, transform shape}},
    every axis plot/.append style={ultra thick}
]
    
    \addplot [
    color=Gray,
    line width=1pt,
    ]
    table[x=roundshow,y=flade,col sep = comma] {iros_data_noniid.csv};



    \addplot [
    color=LimeGreen,
    line width=1pt,
    ]
    table[x=roundshow,y=auade0.15,col sep = comma] {iros_data_noniid.csv};

    \addplot [
    color=Cyan,
    line width=1pt,
    ]
    table[x=roundshow,y=auade0.3,col sep = comma] {iros_data_noniid.csv};

\legend{FLTP,ALFLTP-AU($f_2$=0.15),ALFLTP-AU($f_2$=0.30)}
    
\end{axis}
\end{tikzpicture}
\vspace{-5pt}
\end{subfigure}
\hspace{-20pt}
\begin{subfigure}{0.27\linewidth}
\begin{tikzpicture}
\begin{axis}[
    xlabel={\scriptsize{Round}},
    ylabel={\scriptsize{minFDE}},
    every axis x label/.style={at={(current axis.south)},below=8pt},
    every axis y label/.style={
            at={(ticklabel* cs:0.5)},rotate=90,anchor=center,align=center, above=10pt},
    xmin=0, xmax=250,
    ymin=1.1, ymax=1.4,
    xtick={0,50,100,150,200,250},
    xticklabels={\scriptsize{0},\scriptsize{50},\scriptsize{100},\scriptsize{150},\scriptsize{200},\scriptsize{250}},
    ytick={1.1,1.2,1.3,1.4},
    yticklabels={\scriptsize{1.1},\scriptsize{1.2},\scriptsize{1.3},\scriptsize{1.4}},
    height = 1\linewidth,
    xmajorgrids=true,
    ymajorgrids=true,
    grid style=dashed,
    legend cell align={left},
    legend style={inner sep=0pt,row sep=-3pt},
    legend style={font=\tiny},
    legend style={nodes={scale=0.7, transform shape}},
    every axis plot/.append style={ultra thick}
]
    
    \addplot [
    color=Gray,
    line width=1pt,
    ]
    table[x=roundshow,y=flfde,col sep = comma] {iros_data_noniid.csv};



    \addplot [
    color=LimeGreen,
    line width=1pt,
    ]
    table[x=roundshow,y=aufde0.15,col sep = comma] {iros_data_noniid.csv};
    
    \addplot [
    color=Cyan,
    line width=1pt,
    ]
    table[x=roundshow,y=aufde0.3,col sep = comma] {iros_data_noniid.csv};

\legend{FLTP,ALFLTP-AU($f_2$=0.15),ALFLTP-AU($f_2$=0.30)}
    
\end{axis}
\end{tikzpicture}
\vspace{-5pt}
\end{subfigure}
\hspace{-20pt}
\begin{subfigure}{0.27\linewidth}
\begin{tikzpicture}
\begin{axis}[
    xlabel={\scriptsize{Round}},
    ylabel={\scriptsize{MR}},
    every axis x label/.style={at={(current axis.south)},below=8pt},
    every axis y label/.style={
            at={(ticklabel* cs:0.5)},rotate=90,anchor=center,align=center, above=13pt},
    xmin=0, xmax=250,
    ymin=0.11, ymax=0.15,
    xtick={0,50,100,150,200,250},
    xticklabels={\scriptsize{0},\scriptsize{50},\scriptsize{100},\scriptsize{150},\scriptsize{200},\scriptsize{250}},
    ytick={0.11,0.12,0.13,0.14,0.15},
    yticklabels={\scriptsize{0.11},\scriptsize{0.12},\scriptsize{0.13},\scriptsize{0.14},\scriptsize{0.15}},
    height = 1\linewidth,
    xmajorgrids=true,
    ymajorgrids=true,
    grid style=dashed,
    legend cell align={left},
    legend style={inner sep=0pt,row sep=-3pt},
    legend style={font=\tiny},
    legend style={nodes={scale=0.7, transform shape}},
    every axis plot/.append style={ultra thick}
]
    
    \addplot [
    color=Gray,
    line width=1pt,
    ]
    table[x=roundshow,y=flmr,col sep = comma] {iros_data_noniid.csv};



    \addplot [
    color=LimeGreen,
    line width=1pt,
    ]
    table[x=roundshow,y=aumr0.15,col sep = comma] {iros_data_noniid.csv};

    \addplot [
    color=Cyan,
    line width=1pt,
    ]
    table[x=roundshow,y=aumr0.3,col sep = comma] {iros_data_noniid.csv};

\legend{FLTP,ALFLTP-AU($f_2$=0.15),ALFLTP-AU($f_2$=0.30)}
    
\end{axis}
\end{tikzpicture}
\vspace{-5pt}
\end{subfigure}
\caption{Round-wise comparison between FLTP and ALFLTP-AU}
\label{fig:alfltpau}
\vspace{10pt}
\end{figure*}
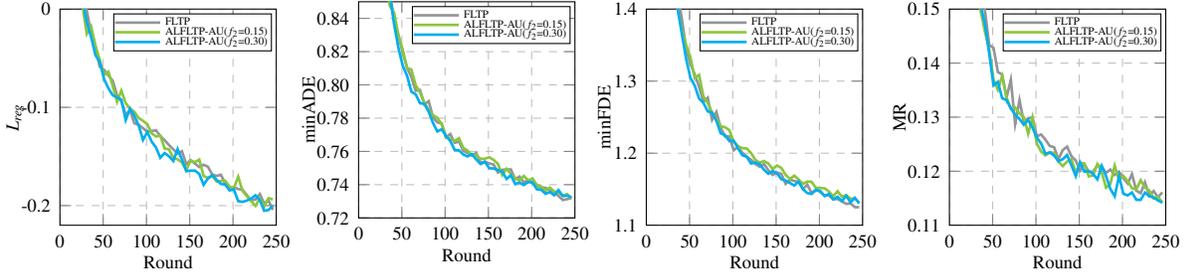

\begin{table}[ht]
\setlength\tabcolsep{2pt}
  \centering
  \renewcommand{\arraystretch}{1.1}
  \begin{tabular}{l|c|c|c|c|c}
    \hline
    Model & Round & NLL($\downarrow$) & minADE($\downarrow$) & minFDE($\downarrow$) & MR($\downarrow$)\\
    \hline
    Centralized HiVT \cite{zhou2022hivt} & - & 0.467 & 0.685 & 1.028 & 0.104\\
    \hline
    Client 0 w/o FL & 50 & 0.897 & 0.992 & 1.732 & 0.207\\
    FLTP & 50 & 0.629 & 0.818 & \textbf{1.318} & 0.140\\ 
    ALFLTP-NLL($f_2$=0.15) & 50 & 0.634 & 0.821 & \textbf{1.318} & \textbf{0.138}\\
    ALFLTP-NLL($f_2$=0.30) & 50 & 0.636 & 0.816 & 1.319 & 0.141\\
    ALFLTP-AU($f_2$=0.15) & 50 & 0.629 & 0.819 & 1.325 & 0.139\\
    ALFLTP-AU($f_2$=0.30) & 50 & \textbf{0.625} & \textbf{0.813} & \textbf{1.318} & 0.141\\
    \hline
    Client 0 w/o FL & 150 & 1.098 & 1.026 & 1.822 & 0.229\\
    FLTP & 150 & \textbf{0.551} & 0.751 & \textbf{1.170} & 0.123\\ 
    ALFLTP-NLL($f_2$=0.15) & 150 & 0.552 & \textbf{0.750} & 1.179 & 0.122\\
    ALFLTP-NLL($f_2$=0.30) & 150 & 0.564 & 0.753 & \textbf{1.170} & 0.120\\
    ALFLTP-AU($f_2$=0.15) & 150 & 0.555 & 0.753 & 1.177 & \textbf{0.119}\\
    ALFLTP-AU($f_2$=0.30) & 150 & 0.554 & 0.752 & \textbf{1.170} & \textbf{0.119}\\
    \hline
    Client 0 w/o FL & 250 & 1.259 & 1.059 & 1.896 & 0.245\\
    FLTP & 250 & 0.527 & 0.730 &  \textbf{1.122} & \textbf{0.114}\\ 
    ALFLTP-NLL($f_2$=0.15) & 250 & 0.532 & 0.732 & 1.139 & 0.116\\
    ALFLTP-NLL($f_2$=0.30) & 250 & 0.536 & 0.733 & 1.126 & \textbf{0.114}\\
    ALFLTP-AU($f_2$=0.15) & 250 & 0.531 & 0.731 & 1.131& 0.115\\
    ALFLTP-AU($f_2$=0.30) & 250 & \textbf{0.526} & \textbf{0.729} & 1.126 & \textbf{0.114}\\
    \hline
  \end{tabular}
  \vspace{10pt}
  \caption{Performance on Argoverse Validation Set. Fraction of clients selected for communication in each round $f_1$ is set to be 0.1.}
  \label{i^*b:roundshotdata}
  \vspace{-5pt}
\end{table}
\subsection{Comparison between FLTP and ALFLTP}
In this section, ALFLTP frameworks using two active client selection metrics together with different degrees of bias are compared with FLTP. From Fig. \ref{fig:alfltpnll} and Fig. \ref{fig:alfltpau}  we can see that:
\begin{itemize}
    \item \textbf{Convergence speed of training loss:} Regression loss of both ALFLTP-NLL and ALFLTP-AU with $f_2=0.30$ converge faster than  with $f_2=0.15$ and FLTP.
    \item \textbf{Round-wise validation performance:} ALFLTP-NLL and ALFLTP-AU with both $f_2=0.15$ and $f_2=0.30$ perform better than FLTP in terms of MR. As MR measures the fraction of scenarios with endpoint errors larger than 2 meters, lower MR demonstrates that ALFLTP-NLL and ALFLTP-AU are more robust to various traffic scenarios in the inference stage.
    \item \textbf{Impact of biased selection-NLL:} After around 100 rounds, FLTP surpasses the performance of ALFLTP-NLL with both $f_2 = 0.15$ and $f_2 = 0.30$ in terms of minADE and minFDE due to biased selection. Notably, ALFLTP-NLL with $f_2 = 0.30$ performs worse than it with $f_2 = 0.15$, because the former introduces more bias.
    \item \textbf{Impact of biased selection-AU:} Compared to ALFLTP-NLL with $f_2=0.15$ and $f_2=0.30$ and ALFLTP-AU with $f_2=0.15$, ALFLTP-AU with $f_2=0.30$ achieves comparable minADE and minFDE to FLTP while does better in terms of MR in most rounds, indicating that a larger $f_2$ helps ALFLTP-AU to find clients with more representative data.
\end{itemize}

 Table \ref{i^*b:roundshotdata} shows detailed global model performance of specific rounds, where we can see that: 
 \begin{itemize}
       \item In the 50th round, ALFLTP-AU with $f_2=0.30$ outperforms other frameworks in NLL, minADE and minFDE.
     \item In the 150th round, ALFLTP-AU with $f_2=0.15$ and $f_2=0.30$ outperform other frameworks in terms of minADE.
      \item In the 250th round, where global models finish training, ALFLTP-AU with $f_2=0.30$ outperforms other frameworks in terms of NLL, minADE and MR.
      \item Although FLTP based HiVT models in the 250th round perform slightly worse than the centralized HiVT, it protects the privacy of the human-driven vehicles by avoiding the data exchange with the server.

 \end{itemize}

In a word, ALFLTP-AU converges faster in regression loss and has better performance in terms of NLL, minADE and MR than FLTP in most rounds. Moreover, ALFLTP-AU shows better and more stable round-wise performance than ALFLTP-NLL.

\section{Conclusion}
In this paper, we propose a privacy-preserving and uncertainty-aware trajectory prediction framework for connected autonomous vehicles using federated learning with a uncertainty-aware global objective. We term this framework as FLTP, where we relax the requirement of collecting raw data of driving scenarios to form a large centralized dataset and let CAVs collect local traffic data and collaboratively train trajectory prediction models without explicit data exchange, thus preserving privacy of traffic participants. We further introduce Active Learning-boosted FLTP (ALFLTP) for client selection in FLTP, where we adopt two uncertainty-aware metrics, negative log-likelihood (NLL) and aleatoric uncertainty (AU) to actively select clients for partial client participation in FLTP. Experiments on Argoverse dataset demonstrate that FLTP significantly outperforms the model trained on local data. In addition, ALFLTP-AU has a faster convergence speed in training regression loss and performs better in terms of NLL, minADE and MR than FLTP in most rounds, and has more stable round-wise performance than ALFLTP-NLL.

\bibliographystyle{unsrt}
\bibliography{references}

\end{document}